\documentclass[11pt]{article}

\usepackage[]{acl}

\usepackage{times}
\usepackage{latexsym}
\usepackage[T1]{fontenc}
\usepackage[utf8]{inputenc}
\usepackage{microtype}
\usepackage{adjustbox}

\usepackage{xcolor}

\newcommand{\mzboito}[1]{\textcolor{blue}{#1}}
\usepackage{amsmath}
\usepackage{amssymb}
\usepackage{amsfonts}
\usepackage{amsbsy}
\usepackage{mathtools}

\usepackage{url}
\usepackage{breakurl}

\usepackage{todonotes}
\usepackage{colortbl}
\usepackage{graphicx,multirow}

\usepackage{booktabs}
\usepackage{enumitem}

\title{ON-TRAC Consortium Systems for the IWSLT 2022\\ Dialect and Low-resource Speech Translation Tasks}

\author{Marcely Zanon Boito$^1$, John Ortega$^2$, Hugo Riguidel$^2$, Antoine Laurent$^2$, \\ \bf Lo{\"i}c Barrault$^2$,
 Fethi Bougares$^3$, Firas Chaabani$^3$, Ha Nguyen$^{1,5}$, \\ \bf Florentin Barbier$^4$,   Souhir Gahbiche$^4$, Yannick Est\`eve$^1$ \\
  $^1$LIA - Avignon University, France,  $^2$LIUM - Le Mans University, France\\
  $^3$ELYADATA - Tunis, Tunisia, $^4$Airbus - France, $^5$LIG - Grenoble Alpes University\\
  \textbf{contact email:} \texttt{yannick.esteve at univ-avignon.fr} 
}
  
\begin{document}
\maketitle
\begin{abstract}
This paper describes the ON-TRAC Consortium translation systems developed for two challenge tracks featured in the Evaluation Campaign of IWSLT 2022: low-resource and dialect speech translation. 
For the Tunisian Arabic-English dataset~(low-resource and dialect tracks), we build an end-to-end model as our joint primary submission, and compare it against cascaded models that leverage a large fine-tuned wav2vec~2.0 model for ASR. Our results show that in our settings pipeline approaches are still very competitive, and that with the use of transfer learning, they can outperform end-to-end models for speech translation (ST).
For the Tamasheq-French dataset~(low-resource track) our primary submission leverages intermediate representations from a wav2vec~2.0 model trained on 234~hours of Tamasheq audio, while our contrastive model uses a French phonetic transcription of the Tamasheq audio as input in a Conformer speech translation architecture jointly trained on automatic speech recognition, ST and machine translation losses. Our results highlight that self-supervised models trained on smaller sets of target data are more effective to low-resource end-to-end ST fine-tuning, compared to large off-the-shelf models. Results also illustrate that even approximate phonetic transcriptions can improve ST scores.
\end{abstract}

\section{Introduction}

The vast majority of speech pipelines are developed for and in \textit{high-resource} languages, a small percentage of languages for which there is a large amount of annotated data freely available~\cite{joshi2020state}. However, the assessment of systems' performance only on high-resource settings can be problematic because it fails to reflect the real-world performance these approaches will have in diverse and smaller datasets.

In this context, the IWSLT~2022~\cite{iwslt:2022} proposes two interesting shared tasks: low-resource and dialect speech translation~(ST). The former aims to assess the exploitability of current translation systems in data scarcity settings. The latter focuses on the assessment of the systems capabilities in \textit{noisy} settings: different 
dialects are mixed in a single dataset of spontaneous speech. For the low-resource task, this year's language pairs are: Tamasheq-French and Tunisian Arabic-English. The latter is also used, in constrained conditions, for the dialect task.

This paper reports the ON-TRAC consortium submissions for the mentioned tasks. 
The ON-TRAC Consortium is composed of researchers from three French academic laboratories, LIA~(Avignon University), LIUM~(Le Mans University) and LIG~(University Grenoble Alpes), together with two industrial partners: Airbus France and ELYDATA.
Our systems for the dialect task focus on the comparison between cascaded and end-to-end approaches for ST. For the low-resource task, we focus on the leveraging of models based on self-supervised learning~(SSL), and on the training of ST models with joint automatic speech recognition~(ASR), machine translation~(MT) and ST losses.

This paper is organized as follows. Section~\ref{sec:related} presents the related work. The experiments with the Tunisian Arabic-English dataset for low-resource and dialect ST tasks are presented in Section~\ref{sec:tuneng}. Results for the Tamasheq-French dataset for the low-resource track are presented in Section~\ref{sec:taqfra}. Section~\ref{sec:conclusion} concludes this work.

\section{Related work}\label{sec:related}

Before the introduction of \textit{direct} or \textit{end-to-end} ST models~\cite{berard2016listen,weiss17_interspeech}, the ST task was 
approached as a \textit{cascaded} problem: the speech is transcribed using an ASR model, and the transcriptions are used to train a classic MT model. The limitations of this approach include the need for extensive transcriptions of the speech signal, and the error propagation between ASR and MT modules. 
In comparison to that, end-to-end ST models propose a simpler encoder-decoder architecture, removing the need for intermediate representations of the speech signal. Although at first, cascaded models were superior in performance compared to end-to-end models, results from recent IWSLT campaigns illustrate how end-to-end models have been closing this gap~\cite{ansari2020findings,bentivogli2021cascade, anastasopoulos-etal-2021-findings}. Moreover, the joint optimization of ASR, MT and ST losses in end-to-end ST models was shown to increase overall performance~\cite{le2020dual,sperber2020consistent}.



SSL models for speech processing are now a popular foundation blocks in speech pipelines~\cite{schneider2019wav2vec, hsu2021hubert, baevski2019effectiveness, baevski2020wav2vec}. These models are large trainable networks with millions, or even billions~\cite{babu2021xls}, of parameters that are trained on unlabeled audio data only.
The goal of training these models is providing a powerful and reusable abstraction block, which is able to process raw audio in a given language or in multilingual settings~\cite{conneau2020unsupervised,babu2021xls}, producing a richer audio representation for the downstream tasks to train with, compared to surface features such as MFCCs or filterbanks. Recent work found considerable performance gains and/or state-of-the-art performance by including these blocks in their target tasks, and more importantly, the final models can be trained with a smaller amount of labeled data, increasing the \textit{accessibility} of current approaches for speech processing~\cite{kawakami-etal-2020-learning,schneider2019wav2vec, hsu2021hubert, baevski2019effectiveness, baevski2020wav2vec}.\footnote{Recent benchmarks for SSL models can be found in \citet{evain21_interspeech,evain2021task,yang21c_interspeech,conneau2022xtreme}.}



\section{Tunisian Arabic-English Experiments}\label{sec:tuneng}

In this section we present our experiments for translating Tunisian Arabic to English in the context of the dialect and low-resource tasks from IWSLT~2022. Section~\ref{sec:tuneng:data} describes the data used in our experiments. 

We investigate two types of ST architectures: end-to-end architectures~(Section~\ref{sec:tuneng:e2est}), and pipeline models~(Section~\ref{sec:tuneng:pipelinest}). For the latter, we include the obtained ASR results. For both, results on the ST tasks are presented in Section~\ref{sec:tuneng:results}.

\subsection{Data}\label{sec:tuneng:data}

The Tunisian Arabic dataset~(LDC2022E01) use in our experiments was developed and provided by LDC\footnote{\url{https://www.ldc.upenn.edu/}} to the IWSLT~2022 participants. It comprises 383\,h of Tunisian conversational speech with manual transcripts, from which 160\,h are also translated into English. Thus, it is a three-way parallel corpus~(audio, transcript, translation). This LDC data consistitutes \textit{basic condition} of the dialect task.
Arabic dialects are the informal form of communication in the everyday life in the Arabic world. Tunisian Arabic is one of several Arabic dialects: there is no standard written Arabic form for this language that is shared by all Tunisian speakers. 
Nevertheless, the transcripts of Tunisian conversations of the LDC2022E01 Tunisian Arabic dataset follow the rules of the Tunisian Arabic CODA – Conventional Orthography for Dialectal Arabic.

For the \textit{dialect adaptation condition}, we use in addition to the LDC2022E01 dataset, the MGB2 dataset~\cite{ali2016mgb}, which is 
composed of 1,200\,h of broadcast news audio recordings in modern standard Arabic~(MSA) from Aljazeera TV programs.
These recordings are associated to captions with no timing information: they
are not verbatims of the speech content, and can be an approximation.
The MGB2 dataset also contains the automatic transcriptions generated by the Qatar Computing Research Institute~(QCRI) ASR system. This external dataset is used for training our ASR systems.

\subsection{Pipeline ST}\label{sec:tuneng:pipelinest}

For our pipeline ST models, we experiment with two different ASR architectures, presented in Section~\ref{sec:tuneng:pipelinest:asr}. We also train two MT models, presented in Section~\ref{sec:tuneng:pipelinest:st}.

\subsubsection{ASR system}\label{sec:tuneng:pipelinest:asr}

\paragraph{End-to-end ASR model.}
Our end-to-end ASR system is implemented on the SpeechBrain toolkit~\cite{speechbrain}. It is composed of a wav2vec~2.0 module, a 1024-dimension dense hidden layer with a Leaky ReLU activation function, and a softmax output layer.
The weights of the wav2vec~2.0 module were initialized from the XLSR-53 model released by Meta~\cite{conneau2020unsupervised}. 
The CTC loss function~\cite{graves2006connectionist} was used during the training process, and two different instances of Adam~\cite{adam} optimizers were used to manage the weight updates: one dedicated to the wav2vec~2.0 module, the other one to the two additional layers. The output of the end-to-end model is based on characters.

The training of our model is separated in two stages. First, we train an end-to-end ASR model in MSA using the MGB2 data. 
To process this data, we used a dictionary of 95 characters~(i.e. 
$95$-dimensional output layer). 
Among the 1,200\,h of speech associated to captions and automatic transcripts in the MGB2 dataset, we keep only 
the audio segments for which the captions and the automatic transcripts are strictly the same. This corresponds to roughly 820\,h of speech.

Once our model in standard Arabic is trained, we use it to initialize our final Tunisian Arabic ASR model. The architecture is kept the same, excluding the $34$-dimensional output layer, and we 
randomly reinitialise the weights of the 2 last layers.
In other words, we keep only 
the weights of the ASR MGB2 fine-tuned wav2vec~2.0 model, 
performing \textit{transfer learning} from MSA to Tunisian Arabic.
We then train the end-to-end ASR model on the Tunisian audio data of the LDC2022E01 dataset and its normalized transcription.
Lastly, we train a 5-gram language model~(LM) on the normalized transcriptions.

\paragraph{Hybrid HMM/TDNN ASR system.}
In addition to the end-to-end ASR system describe above, we train a Kaldi-based system~\cite{povey2011kaldi}. 
The acoustic model uses chain models with the TDNN architecture and 40-dimensional high-resolution MFCCs extracted from frames of 25\,ms length and 10\,ms shift, applying usual data augmentation methods: speed perturbation at rates of $0.9$, $1.0$, and $1.1$, and 
spectral
augmentation. We employ a graphemic lexicon of 88k words, and we use a 3-gram LM built using
the \textit{SRILM} toolkit~\cite{stolcke2002srilm} with the Kneser-Ney smoothing. This 3-gram LM is trained using the transcripts of the training set and the vocabulary covering all the words of the graphemic lexicon.
 

\begin{table}
\begin{center}
\begin{tabular}{llcc}
    \toprule
    System & Description &  valid & test \\
    \midrule
    primary & E2E w/o LM & 41.1  & 45.1   \\
    not submitted & HMM/TDNN & 50.3 & - \\ 
    post-evaluation & E2E + 5-gram  & 38.3 & 41.5  \\ 
    \bottomrule
\end{tabular}
\end{center}
\caption{Results for Tunisian Arabic ASR systems in terms of WER. Submissions to the low-resource track.}
\label{tab:tna-en:asr}
\end{table}

\paragraph{ASR performance.}
Tunisian Arabic ASR results for 3 different models are presented in Table~\ref{tab:tna-en:asr}. 
The primary system is the end-to-end ASR model described above, without LM rescoring.
The second row presents the result for the hybrid HMM/TDNN system.
Due to its lower performance on the validation data in comparison to the end-to-end system, we decided to not submit this system. 
The last row presents the results for the end-to-end ASR with the 5-gram LM, 
a post-evaluation result.

\subsubsection{MT model}\label{sec:tuneng:pipelinest:st}


We train two MT models 
using the \textit{fairseq} toolkit~\cite{ott2019fairseq}. The first model~(\textbf{contrastive1}) is an bi-LSTM model from \citet{luong2015effective}, trained using the \texttt{lstm\_luong\_wmt\_en\_de} recipe\footnote{\url{https://fairseq.readthedocs.io/en/latest/_modules/fairseq/models/lstm.html}}. Both encoder and decoder consists of 4 LSTM layers, and the input 
is at the sub-word level using a BPE vocabulary of 8,000 units, 
trained on the target language.

The second model~(\textbf{contrastive2}) is a fully convolutional model following the \texttt{fconv\_wmt\_en\_fr}\footnote{\url{https://fairseq.readthedocs.io/en/latest/models.html}} sequence-to-sequence architecture from \citet{DBLP:journals/corr/GehringAGYD17}. It consists of 15 encoder and decoder layers, working on the 
sub-word level with input and output vocabularies of 4,000 BPE units.

\subsection{End-to-end ST}\label{sec:tuneng:e2est}
The end-to-end ST model is a Conformer model~\cite{gulati2020conformer} based on the \textit{Espnet} toolkit~\cite{watanabe2018espnet}. This system is trained using 80-channel log-mel filterbank 
features computed on a 25\,ms window with a 10\,ms shift. We also use speed perturbation at ratio 
$0.9$, $1.0$, $1.1$ and \textit{SpecAugment}~\cite{park2019specaugment} with 2 frequency masks and 5 time masks. In addition, a global Cepstral Mean and Variance Normalization~(CMVN) technique is applied on the top of our features. 

Our Conformer model consists of a 6-block 
Conformer encoder and a 6-block Transformer decoder. We use 1,000 BPE as the modeling 
units. The model is trained for 100 epochs and the last 10 best checkpoints are averaged to create the final model.

\subsection{Results}\label{sec:tuneng:results}

\begin{table}
\begin{center}
\resizebox{\columnwidth}{!}{
\begin{tabular}{lclcc}
    \toprule
    System & Track & Description & valid & test \\
    \midrule
    primary & LR/D & End-to-end & 12.2  & 12.4   \\
    contrastive1 & LR & Cascade & 15.1 & 13.6 \\ 
    contrastive2 & LR & Cascade & 12.8 & 11.3  \\ 
    post-evaluation & LR & Cascade & 16.0 & 14.4  \\ 
    \bottomrule
\end{tabular}}
\end{center}
\caption{Results for Tunisian Arabic to English translation systems in terms of \%BLEU for low-resource~(LR) and dialect~(D) tracks.}
\label{tab:tna-en:st}
\end{table}

Table~\ref{tab:tna-en:st} presents our ST results for dialect and low-resource tracks. Our primary system for both tracks is the end-to-end system presented in Section~\ref{sec:tuneng:e2est}. The two pipeline systems, \textit{contrastive1} and \textit{contrastive2}, are composed by the end-to-end ASR model, and they vary on the MT model used~(presented in Section~\ref{sec:tuneng:pipelinest:st}). Since ASR models use external data~(MGB2), these submissions are for the low-resource track only. Finally, the \textit{post-evaluation} model is the composition of the \textit{post-evaluation} end-to-end ASR model from Section~\ref{sec:tuneng:pipelinest:asr}, and the MT model from \textit{contrastive1}. 

We observe that our cascaded models are very competitive compared against our end-to-end model~(primary submission): our best ST result is 
obtained using the \textit{contrastive1}. The \textit{post-evaluation} model, which adds an 5-gram LM on the end-to-end ASR module, achieves even better scores.
We believe that part of the reason this model is 
effective is the addition of the data in MSA from the MGB2 dataset, that is used to pre-train the end-to-end ASR model. Thus, the comparison between our cascaded and end-to-end models is not exactly fair, as out end-to-end model is trained on less data.

Moreover, we would like to highlight that although this dataset is offered as part of the \textit{low-resource} track, we do not consider this setting to be one of data scarcity: 160\,h of translated speech are available. We do, however, find this dataset to be extremely complex to work with. 
That is because there are multiple regional dialects from Tunisia mixed in the data, which makes the ST task harder.
These regional dialects differ mainly on their accent, but  sometimes also in terms of 
vocabulary and expression.

Nonetheless, we find that the real challenge for processing 
this data comes from its nature.
This dataset is a collection of telephonic conversations, where the acoustic conditions can be sometimes very challenging: some phone calls are made from mobile phones in very noisy environments, and sometimes some portions of audio recordings are saturated because of 
sudden high audio input gain.

By computing the WER on each audio recording in the validation set using our best ASR model, 
we observe that the lowest one achieved is 18.3\%, while the highest one is 88.5\%.
Thus, we achieve a global WER of 38.3\%~(\textit{post-evaluation} in Table~\ref{tab:tna-en:asr}), with a standard deviation is 12.3\%. This illustrates the high variability in terms of audio quality that might exist in this dataset.

\section{Tamasheq-French Experiments}\label{sec:taqfra}

In this section we present our experiments for the Tamasheq-French dataset in the context of the low-resource ST track. This dataset, recently introduced in \citet{boito2022speech}, contains 17\,h of speech in the Tamasheq language, which corresponds to 5,829~utterances translated to French. 
Additional audio data was also made available through the \textit{Niger-Mali audio collection}: 224\,h in Tamasheq and 417\,h in geographically close languages (French from Niger, Fulfulde, Hausa, and Zarma).\footnote{\url{https://demo-lia.univ-avignon.fr/studios-tamani-kalangou/}} For all this data, the speech style is radio broadcasting, and the dataset 
presents no transcription.


Our experiments are separated in two different investigation branches:
\begin{enumerate}
    \item The exploitation of SSL wav2vec~2.0 models~\cite{baevski2020wav2vec} for low-resource direct speech-to-text translation;
    \item  The production of \textit{approximate} phonetic transcriptions for attenuating the challenge of training in low-resource settings.
\end{enumerate}
We start by presenting the models proposed for the first branch: the SSL models pre-trained and/or fine-tuned for Tamasheq in Section~\ref{sec:taqfra:ssl}, the \textit{pipeline} experiments that use wav2vec~2.0 models as feature extractors in Section~\ref{sec:taqfra:pipeline}, and our primary system, an end-to-end architecture that directly fine-tunes a wav2vec~2.0 model, in Section~\ref{sec:taqfra:end2end}. Section~\ref{sec:taqfra:asrst} focuses on the second branch of experiments, presenting our contrastive model that is based on the joint optimization of ASR, MT and ST losses. This is made possible by the use of a French ASR system for generating an approximated phonetic transcription of the Tamasheq audio. In Section~\ref{sec:taqfra:discussion}, we present and discuss our results, and lastly, Section~\ref{sec:taqfra:other} describes some less-successful experiments.

\subsection{SSL models}\label{sec:taqfra:ssl}

\paragraph{Pre-trained models.}
We train two wav2vec~2.0 \textit{base} models
using the Niger-Mali audio collection. The \textit{Tamasheq-only} model uses the 224\,h in Tamasheq, and the \textit{Niger-Mali} model uses all the data available: 641\,h in five languages. Additionally, we include in the training data for both models the 19\,h present in the \textit{full} release of the Tamasheq-French corpus.\footnote{\url{https://github.com/mzboito/IWSLT2022_Tamasheq_data}} Therefore, both models are pre-trained on the target data. 
For training them, we use the same hyperparameters from the original wav2vec~2.0, as well as the original 
\textit{fairseq}~\cite{ott2019fairseq} implementation. These models are trained until $500$k updates on $16$ Nvidia Tesla V100~(32GB), and they are available for download at HuggingFace.\footnote{\url{https://huggingface.co/LIA-AvignonUniversity}} 

\noindent \textbf{Fine-tuned models.} We experiment with the 7K large French wav2vec~2.0 model~(LB-FR-7K) from the \textit{LeBenchmark}~\cite{evain21_interspeech}, and the multilingual XLSR-53~\cite{conneau2020unsupervised}. Both models are fine-tuned on the 243\,h of Tamasheq~($224$\,h $+ 19$\,h) for approximately $20$k updates on $4$ Nvidia Tesla V100~(32GB). \mzboito{} 
Finally, using the Tamasheq-only model, we also experiment fine-tuning it for the ASR task in MSA~(primary ASR model from Section~\ref{sec:tuneng:pipelinest}). 

\subsection{Pipeline SSL+ST models}\label{sec:taqfra:pipeline}
Our models are very close to the recipe for low-resource ST from wav2vec~2.0 features described in \citet{evain2021task}. 
We use the \textit{fairseq s2t} toolkit~\cite{wang2020fairseq} for training an end-to-end ST Transformer model~\cite{vaswani2017attention} with 4 heads, dimensionality of 256, inner projection of 1,024, 6 encoder and 3 encoder layers.
The Transformer is preceded by a 1D convolutional layer~(k=5, stride=2) for down-projecting the wav2vec~2.0 large~(1,024) or base~(768) features into the Transformer input dimensionality.
These models are trained for 500 epochs using the Adam optimizer~\cite{adam} with 10k warm-up steps. For decoding, we use beam search with a beam size of 5. For these models and the ones from Section~\ref{sec:taqfra:end2end}, we generate a 1k unigram vocabulary for the French text using \textit{Sentencepiece}~\cite{kudo2018sentencepiece}, with no pre-tokenization.

Lastly, we include baseline results that replace wav2vec~2.0 features by 80-dimensional mel filterbank~(MFB) features. In this setting, the CNN preceding the transformer encoder is identical from the one in \citet{evain2021task}.

\subsection{End-to-end SSL+ST models}\label{sec:taqfra:end2end}
Training an end-to-end ST model from a pre-trained speech encoder was first proposed in \citet{li-etal-2021-multilingual}.
In this work, our end-to-end ST model is similar to the end-to-end ASR model presented in Section~\ref{sec:tuneng:pipelinest:asr}. It is also implemented on \textit{SpeechBrain}, and it comprises a wav2vec~2.0 as speech encoder, followed by a linear projection, and the Transformer Decoder from Section~\ref{sec:taqfra:pipeline}. The weights for the wav2vec~2.0 speech encoder are initialized from one of the models in Section~\ref{sec:taqfra:pipeline}, and the model is trained on the NLL 
loss. As in Section~\ref{sec:tuneng:pipelinest}, two different instances of the Adam optimizer manage the weight updates: one dedicated to the wav2vec~2.0 module, the other one to the following layers. 

Inspired by the layer-wise investigation for wav2vec~2.0 models described in \citet{pasad2021layer}, we explore reducing the number of layers in the Transformer encoder that is internal to the wav2vec~2.0 module. This is based on their finding that the Transformer encoder behaves in an auto-encoder fashion and therefore, the intermediate representations might contain a higher level of abstraction from the speech signal. In their work, they show that re-initializing the weights of the final Transformer Encoder layers increases performance in ASR fine-tuning.

Different from that, we propose to remove these layers altogether, which we believe is beneficial for low-resource ST fine-tuning for two reasons. First, a reduced wav2vec~2.0 module will still have considerable capacity for encoding the speech, and second, this reduction in number of trainable parameters might facilitate training. 

For implementing this 
model, we simply drop the $N$ final encoder layers from our training graph, keeping the final projection. 
We refer to this architecture as \textit{W2V-N+ST}, where $N$ is the number of layers, starting from the first, kept during ST training. 


\subsection{End-to-end ASR+ST models}\label{sec:taqfra:asrst}

We investigate a ST architecture that jointly optimizes ST, MT and ASR losses, as in \citet{le2020dual}. For this evaluation campaign however, no Tamasheq transcript nor phonetic transcription was provided, so we create an approximate phonetic transcription~(Section~\ref{sec:taqfra:asrst:asr}) that we use in our end-to-end joint system for ST~(Section~\ref{sec:taqfra:asrst:st}).


\subsubsection{Phonetic transcription for Tamasheq}\label{sec:taqfra:asrst:asr}
The Tamasheq is a Tuareg language spoken by around 500  thousand speakers, mainly from northern Mali.
Its phonological system contains 5 vowels~(+2 short vowels) and approximately 21 consonants if we ignore the 6 consonants of Arabic origin that are of marginal use~(mostly for loanwords)~\cite{heath2005}. This leads to a set of 26 phonemes.  
Almost all of those phonemes appear to occur in French, which contains 36 phonemes, 16 vowels, 17 consonants and 3 glides.


This motivates to use a phonetizer pretrained on French in order to ``transcribe'' the Tamasheq signal into a sequence of pseudo-Tamasheq phonemes.
A phonetic force alignment using a pre-trained Kaldi~\cite{povey2011kaldi} chain-TDNN acoustic model was used, followed by an ASR system trained using ESPNet~\cite{watanabe2018espnet}. The model is trained on MFB 
features, and it uses 12 blocks of Conformer~\cite{gulati2020conformer} encoders, followed by 6 blocks of Transformer decoders. It uses a hybrid loss between attention mechanism and CTC~\cite{graves2006connectionist}.

The French corpus is composed of approximately 200\,h coming from ESTER1\&2~\cite{galliano2009ester}, REPERE~\cite{giraudel2012repere} and VERA~\cite{goryainova2014morpho}. No LM was used, and the phoneme error rate 
achieved on the ESTER2 test corpus is of 7,7\%~(silences are not ignored).

We highlight that there is no simple automatic way to evaluate the quality of the phonetic transcriptions  we generated on Tamasheq. We however, manually verified some transcriptions and confirmed that they seemed to be of overall good quality.

\subsubsection{Architecture}\label{sec:taqfra:asrst:st}

The system is based on the \textit{ESPNet2}~\cite{inaguma-etal-2020-espnet} ST recipe.\footnote{\url{https://github.com/espnet/espnet/tree/master/espnet2/st}}
This end-to-end model is made 
of 12 blocks of conformer encoders~(hidden size of dimension $1024$), followed by 3 blocks of transformer decoders~(
hidden size of dimension $2048$). Input features are 512-dimensional MFB features extracted from the 
wave signal.

Three losses are jointly used for training, as described in Equation~\ref{eq1}. There, $\mathcal{L}_{ST}$ is the loss for Tamasheq speech to French text translation; $\mathcal{L}_{MT}$ is the loss for Tamasheq pseudo-phonetic transcription to French text translation; and  $\mathcal{L}_{ASR}$ is the loss for Tamasheq speech to Tamasheq pseudo-phonetic transcription.  
\begin{equation}
    \mathcal{L} = 0.3 \times \mathcal{L}_{ST} + 0.5 \times \mathcal{L}_{MT} + 0.2 \times \mathcal{L}_{ASR}
    \label{eq1}
\end{equation}


\subsection{Results}\label{sec:taqfra:discussion}

\begin{table}
\begin{center}
\resizebox{\columnwidth}{!}{
\begin{tabular}{lccc}
    \toprule
     System & Description & valid & test \\
    \midrule
    \textbf{primary} & E2E, W2V-6+ST & 8.34  &  5.70 \\ 
    \textbf{contrastive} & E2E, ASR+ST  & 6.40 & 5.04 \\ 
    \midrule 
    contrastive2 & pipeline, W2V-ASR+ST  & 3.62 & 3.17 \\
    contrastive3 & pipeline, W2V-FT+ST  & 2.94 & 2.57 \\
    baseline & pipeline & 2.22 & 1.80 \\
    \bottomrule
\end{tabular}}
\end{center}
\caption{Results for the pipeline and end-to-end~(E2E) Tamasheq-French ST systems in terms of \%BLEU score. 
The first two rows present our submitted systems, while the reminder are complementary post-evaluation results.}
\label{tab:results:tmh-fra}
\end{table}

Results are presented in Table~\ref{tab:results:tmh-fra}. Our primary submission~(W2V-6+ST) 
uses the Tamasheq-only wav2vec~2.0 base model,  with only 6 transformer encoder layers~(from a total of 12). Results with different numbers of layers are present in the Appendix~\ref{app:taqfra:intermediatee2e}. 
Our contrastive submission is the end-to-end model from Section~\ref{sec:taqfra:asrst}. Finally, the three last rows present complementary results, including a baseline trained on MFB features, and two pipeline models. The \textit{contrastive2} uses the Tamasheq-only wav2vec~2.0 model fine-tuned for the Arabic ASR task from Section~\ref{sec:tuneng:pipelinest} as feature extractor,  while \textit{contrastive3} extracts features from the Niger-Mali wav2vec~2.0 base model fine-tuned on Tamasheq. Other pipeline SSL+ST models achieved lower scores, and their results are grouped in Appendix~\ref{app:taqfra:pipeline}.

Looking at our results, and concentrating on SSL models, we notice that models that use wav2vec~2.0 
as feature extractor~(\textit{contrastive2} and \textit{contrastive3}) achieve better performance compared to a baseline using MFB features. However, this finding does not hold for the wav2vec~2.0 large models fine-tuned on Tamasheq~(XLSR-53 and LB-FR-7K), which scored as poorly as our baseline~(results in Appendix~\ref{app:taqfra:pipeline}). 
We find this result surprising, specially in the case of the multilingual model~(XLSR-53). This could mean that these large models are not useful as feature extractors for low-resource settings, even after task-agnostic fine-tuning on the target language.

Regarding the fine-tuning procedure, as in \citet{evain2021task}, we notice that ASR fine-tuning is more beneficial to ST than task-agnostic fine-tuning: \textit{contrastive2} achieves better scores compared to \textit{contrastive3}. We find this result interesting, considering that the ASR fine-tuning performed in this case did not targeted Tamasheq, but MSA. This could mean that, when languages are sufficiently similar, ASR fine-tuning in a different language could be performed for increasing the performance on a low-resource language without transcripts.

Regarding our primary system, we found better results by reducing the amount of trainable encoder layers inside the wav2vec~2.0 module. We also investigated freezing it partially or entirely 
during end-to-end ST training, but this resulted in performance decrease in the validation set. 

Regarding the different wav2vec~2.0 models trained~(Section~\ref{sec:taqfra:ssl}), and focusing on 
our primary model, we find that similar to pipeline SSL+ST models, we achieved our best results with base architectures~(Tamasheq-only and Niger-Mali). 
Close seconds to the performance obtained with our primary model~(on the validation set) were the models using the same wav2vec~2.0 modules from \textit{contrastive2} and \textit{contrastive3}.

These results indicate that having a dedicated wav2vec~2.0 model trained on the target or on close languages is indeed better than fine-tuning large monolingual~(LB-FR-7K) or multilingual~(XLSR-53) models.\footnote{By \textit{close} we mean: (1)~languages that are geographically close and with a known degree of lexical borrowing; (2)~similar speech style and recording settings.} This is particularly interesting considering that the Tamasheq-only model is trained with only 234\,h of speech, whereas XLSR-53 learned from approximately 56 thousand of hours. We believe that more investigation is necessary in order to confirm the observed trend. 
Finally, we find the gap between the primary's performance in validation and test sets surprising, and we intend to investigate this further as well.

Concluding, the \textit{contrastive} model we propose in our submission presents a different approach for low-resource ST. By creating an approximate transcription of the Tamasheq audio, we are able to train more effectively, reaching a performance close to our primary model for the test set. This illustrates how transcriptions can be an effective form of increasing performance in low-resource settings, even when these are automatically generated. A possible extension of this work would be the combination of our primary and contrastive models: by inserting the primary's wav2vec~2.0 speech encoder into the training framework from the contrastive model, one can hypothesize that we could achieve even better scores.

\subsection{Other Approaches}\label{sec:taqfra:other}

\paragraph{XLS-R ST model.}
During development, we tried 
to apply XLS-R for translation~\cite{babu2021xls}, using 
the implementation available on the HuggingFace.
\footnote{\url{https://huggingface.co/facebook/wav2vec2-xls-r-300m-21-to-en}} In this approach, we aimed to use the pre-trained model, that is trained on 21 source languages with one target language~(English), called \textit{wav2vec2-xls-r-300m-21-to-en} to first translate the Tamasheq validation set to English. Then, as a second step, to translate the English system output to French. However, we observed 
that the decoder, based on a mBART~\cite{liu2020multilingual}, repeated several groups of tokens during decoding of up to hundreds of times. For example, the phrase: ``the sun was shining in the sky'' for the sentence: ``In the evening, the sun was shining in the sky, and the sun was shining in the sky...'' was repeated 32 times. This illustrates that out-of-shelf models can still fail to provide decent results in zero-shot settings.

\paragraph{ST fine-tuning for large wav2vec~2.0 models.}
All end-to-end models described in Section~\ref{sec:taqfra:end2end} are trained on a single Nvidia Tesla V100~(32GB). This limited our investigation using large wav2vec~2.0 models, since these 
only fit in this size of GPU after extreme reduction of the decoder network. Therefore, we find difficult to assess if the inferior performance of these large end-to-end models is due to the architecture size, or due to the speech representation produced by the wav2vec~2.0 models. In any case, reducing the number of encoder layers, and freezing some of the initial ones, resulted in better performance. The attained scores were however still inferior compared to pipeline models.

\section{Conclusion}\label{sec:conclusion}

In this paper we presented our results for two IWSLT 2022 tasks: dialect and low-resource ST. Focusing on the Tunisian Arabic-English dataset~(dialect and low-resource tasks), 
we trained an end-to-end ST model as primary submission for both tasks, and contrastive cascaded models that used external data in MSA for the low-resource track. Our cascaded models turned out to outperform slightly our end-to-end model, which we believe might be due to the additional 820\,h of data in MSA that was used to pre-train our end-to-end ASR model. Finally, we observe a considerable variability in our ASR results, hinting that the quality of this dataset might be mixed.

Our experiments with the Tamasheq-French dataset~(low-resource task) included the training and application of wav2vec~2.0 models for ST as either feature extractors or speech encoders. We find the latter to be more beneficial: by fine-tuning half of a wav2vec~2.0 base model trained on the Tamasheq language on the ST task, we achieve our best results. Between our findings regarding the use of SSL models for low-resource ST, we highlight two interesting points: first, we find that fine-tuning wav2vec~2.0 models for the ASR task turns out to be effective even when the fine-tuning and target languages are not the same. Second, we disappointingly observe that large models perform poorly in this low-resource setting, even after fine-tuning in the target language. These last results hint that it might be more beneficial to train wav2vec~2.0 in smaller sets of unlabeled target data~(or in related languages in the same speech settings) than fine-tuning massive off-the-shelf SSL models.

Concluding, we also investigated the generation of approximate transcriptions on Tamasheq by using a French ASR model. Using these transcriptions to jointly constrain an end-to-end ST model on ASR, MT and ST losses, we achieved our second best reported results. This illustrates that even automatically generated approximate transcriptions can reduce the challenge of performing ST in low-resource settings.

\section*{Acknowledgements}
This work was funded by the French Research Agency (ANR) through the ON-TRAC project under contract number ANR-18-CE23-0021. It was also partially funded by the European Commission through the SELMA project under grant number 957017. It used HPC resources from GENCI-IDRIS: grants 2020-A0111012991, 2021-AD011013317, 2021-AD011013331 and 2021-AD011012527. The authors would like to thank Daniel Luzzati from LIUM for his help on the Tamasheq phonological system.

\bibliography{anthology,custom}

\clearpage
\appendix

\section{Tamasheq-French Experiments}

\subsection{ST fine-tuning from intermediate layers}\label{app:taqfra:intermediatee2e}

\begin{table}[h!]
\centering
\begin{tabular}{ccc}\toprule
\# layers & \multicolumn{1}{c}{valid}         & \multicolumn{1}{c}{test}         \\\midrule
12 (all)        & 3.68 &                2.34 \\
11        &    4.40   &  3.21  \\
10        &   5.96   &  4.11 \\
9         &   7.32   &  5.40 \\
8         & 7.64 & 5.64 \\
7         & 8.29 & \textbf{6.00} \\
6         & \textbf{8.34} & 5.70 \\
5         & 7.88 & 5.13 \\
4         & 6.54 & 4.02 \\\bottomrule
\end{tabular}
\caption{Post-evaluation results for the end-to-end W2V-N+ST models from Section~\ref{sec:taqfra:end2end}, using different $N$ values~(number of layers). All models were trained using the Tamasheq-only wav2vec~2.0 base model. Best results in bold.}
\label{app:taqfra:e2e:tab}
\end{table}

\subsection{Pipeline SSL+ST Results}\label{app:taqfra:pipeline}

\begin{table}[h!]
\center
\resizebox{\columnwidth}{!}{
\begin{tabular}{lccc}
\toprule
W2V model     & Fine-tuning   & valid & test \\
\midrule
LB-FR-7K      & -             & 2.36  & 1.80 \\
LB-FR-7K      & Task-agnostic & 2.48  & 1.92 \\
XLSR-53       & -             & 2.05  & 1.42 \\
XLSR-53       & Task-agnostic & 1.99  & 1.91 \\
Tamasheq-only & -             & 2.99  & 2.42 \\
Tamasheq-only & ASR (Arabic)  & \textbf{3.62}  & \textbf{3.17} \\
Niger-Mali    & -             & 2.81  & 2.68 \\
Niger-Mali    & Task-agnostic & 2.94  & 2.57 \\ \bottomrule
\end{tabular}}
\caption{Post-evaluation results for the pipeline SSL+ST models from Section~\ref{sec:taqfra:pipeline}. Task-agnostic corresponds to the fine-tuning on 243\,h of Tamasheq, as described in Section~\ref{sec:taqfra:ssl}. Best results in bold.}
\label{app:taqfra:pipeline:results}
\end{table}

\end{document}